\documentclass[conference]{IEEEtran}
\IEEEoverridecommandlockouts
\usepackage{cite}
\usepackage{amsmath,amssymb,amsfonts}
\usepackage{algorithmic}
\usepackage{graphicx}
\usepackage{textcomp}
\usepackage{xcolor}
\usepackage{tabularx}

\usepackage{graphicx}      % 用于缩放表格
\usepackage{booktabs}      % 提供更美观的表格线
\usepackage{multirow}      % 用于合并行
\usepackage{makecell}      % 用于优化单元格内容
\usepackage{pifont}

\usepackage[hidelinks]{hyperref}
\usepackage{multirow}      % 用于合并行
\usepackage{tabularx}      % 用于自适应列宽
\usepackage{makecell}      % 用于在单元格内换行
\newcommand{\cmark}{\ding{51}} % 对勾
\newcommand{\xmark}{\ding{55}} % 叉号

\begin{document}

\title{Dialogue Director: Bridging the Gap in Dialogue Visualization for Multimodal Storytelling}

\DeclareRobustCommand*{\IEEEauthorrefmark}[1]{%
    \raisebox{0pt}[0pt][0pt]{\textsuperscript{\footnotesize\ensuremath{#1}}}}
\author{\IEEEauthorblockN{Min Zhang\IEEEauthorrefmark{1}\IEEEauthorrefmark{,}\IEEEauthorrefmark{2},
Zilin Wang\IEEEauthorrefmark{1}\IEEEauthorrefmark{,}\IEEEauthorrefmark{2},
Liyan Chen\IEEEauthorrefmark{1}\IEEEauthorrefmark{,}\IEEEauthorrefmark{2}*,
Kunhong Liu\IEEEauthorrefmark{1}\IEEEauthorrefmark{,}\IEEEauthorrefmark{2}*,
Juncong Lin\IEEEauthorrefmark{3},
}
\IEEEauthorblockA{\IEEEauthorrefmark{1}School of Film, Xiamen University, Xiamen, China}
\IEEEauthorblockA{\IEEEauthorrefmark{2}Key laboratory of Digital Protection and Intelligent Processing of lntangible Cultural Heritage \\
of Fujian and Taiwan, Ministry of Culture and Tourism, China}
\IEEEauthorblockA{\IEEEauthorrefmark{3}School of Informatics, Xiamen University, China
}
\IEEEauthorblockA{\{andysummer0715, wanhhe\}@stu.xmu.edu.cn, \{chenliyan, lkhqz,  jclin\}@xmu.edu.cn}

\thanks{* denotes corresponding authors. This work is supported by the Natural Science Foundation of China (No.62077039) and the public technology service platform project of Xiamen City (No.3502Z20231043).}
}

\maketitle

\begin{abstract}
Recent advances in AI-driven storytelling have enhanced video generation and story visualization. However, translating dialogue-centric scripts into coherent storyboards remains a significant challenge due to limited script detail, inadequate physical context understanding, and the complexity of integrating cinematic principles. To address these challenges, we propose Dialogue Visualization, a novel task that transforms dialogue scripts into dynamic, multi-view storyboards.
We introduce Dialogue Director, a training-free multimodal framework comprising a Script Director, Cinematographer, and Storyboard Maker. This framework leverages large multimodal models and diffusion-based architectures, employing techniques such as Chain-of-Thought reasoning, Retrieval-Augmented Generation, and multi-view synthesis to improve script understanding, physical context comprehension, and cinematic knowledge integration.
Experimental results demonstrate that Dialogue Director outperforms state-of-the-art methods in script interpretation, physical world understanding, and cinematic principle application, significantly advancing the quality and controllability of dialogue-based story visualization.
\end{abstract}


\section{Introduction} \label{sec:intro}
AI-driven video generation has revolutionized automated film production, but creating coherent long-form videos with frequent shot transitions remains challenging~\cite{videoworldsimulators2024,10688206,10645369,10687813}. Montage, a key element in filmmaking~\cite{reisz_millar_al_2010}, demands consistency in characters and scenes across shots. Advances in story visualization~\cite{li2018storygan} show promise by generating image sequences from text while addressing scene transitions and character consistency. However, existing methods~\cite{CPCSVTS10.1007/978-3-030-58520-4_2, TSli-lukasiewicz-2022-learning, VLC-StoryGANmaharana2021integrating} face high training costs, reliance on large labeled datasets, and challenges in aligning visual fidelity with textual descriptions due to information loss.

\begin{figure}[t]
  \centering
  \includegraphics[width=1.0\linewidth]{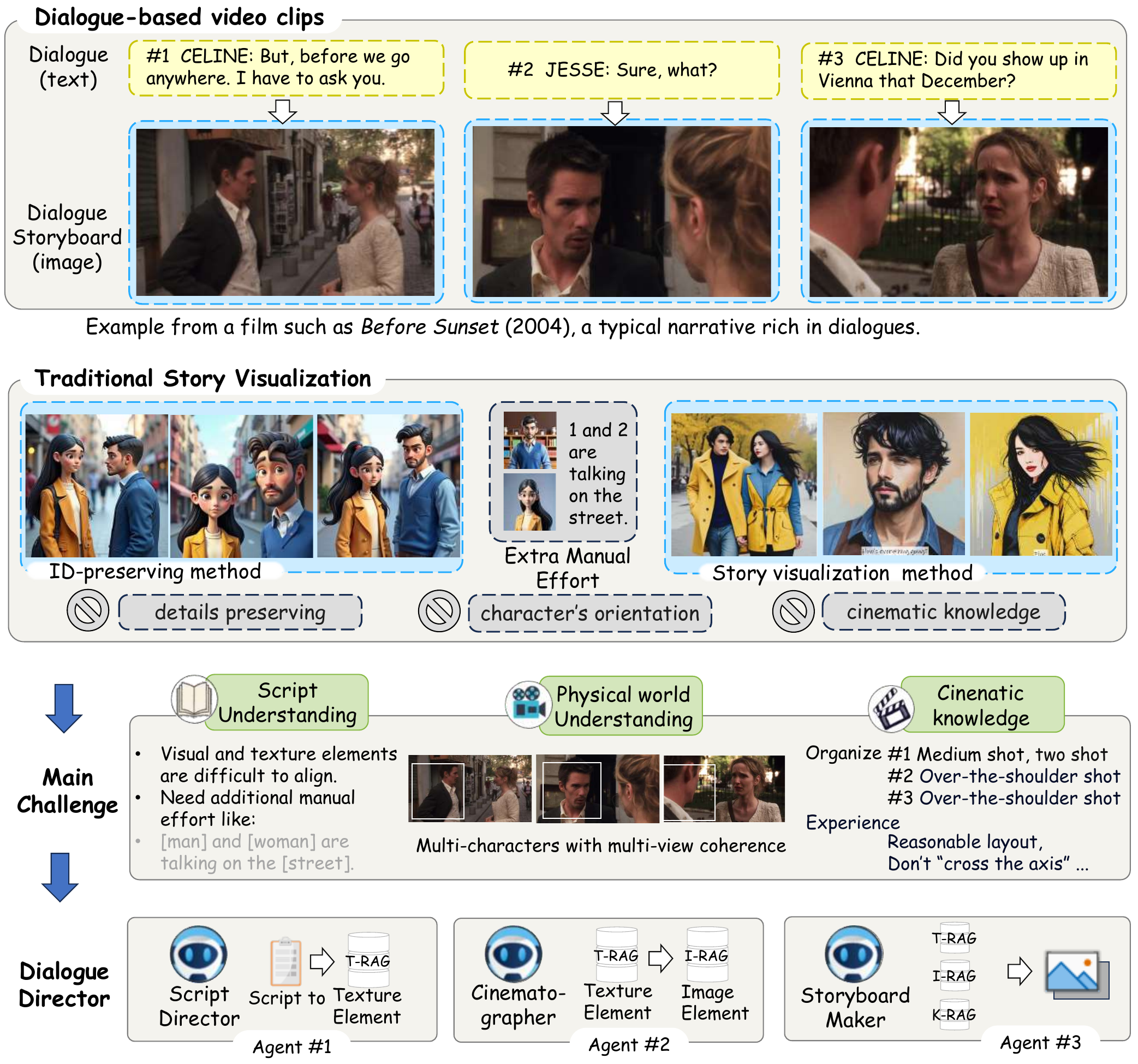}
    \vspace{-0.5cm}
  \caption{Motivation of Dialogue Visualization and our framework. 
  Traditional story visualization methods meet with challenges when handling dialogues, while Dialogues play a significant role in storytelling. Existing methods may be good at ID-preserving, but rely on extra manual effort, perform not well in details preserving, character’s orientation, and cinematic knowledge. }
  \label{1}
  \vspace{-0.5cm}
\end{figure}

Advances in controllable diffusion \cite{CDxiao2024omnigen, wang2024ensembling, wang2024v,iclight} and large language models (LLMs) \cite{LLMmiao2023efficientgenerativelargelanguage} have mitigated some of these challenges by reducing training costs and enabling coherent character generation even in few- or zero-shot scenarios \cite{huang2024resolvingmulticonditionconfusionfinetuningfree, CDxiao2024omnigen, zhou2024storymakerholisticconsistentcharacters, zhou2024storydiffusionconsistentselfattentionlongrange}. However, as shown in \figurename~\ref{1}, they rely on extra manual effort to translate textual cues (may be implicit in dialogue) into visual information for better understanding of visualization models, failing in detail-preserving, character's orientation-controlling and principle of cinema integration. In consequence, dialogue visualization remains uniquely challenging due to sparse descriptive details in dialogue-driven narratives. Generating coherent visuals requires a nuanced understanding of physical scenes, character relationships, and cinematic conventions, including continuity, perspective, and techniques like reasonable layout and axis-crossing \cite{reisz_millar_al_2010}. Integrating cinematic principles into generative models is especially non-trivial, necessitating a framework that bridges textual descriptions, scene layouts, and filmmaking techniques.

Dialogue visualization presents three challenges: (1) aligning sparse, multimodal cues in dialogue scripts with corresponding visuals, (2) maintaining sequence coherence while ensuring realistic multi-view character representation, and (3) applying cinematic principles to structure storyboards that adhere to professional standards. 
Addressing these challenges needs a paradigm capable of extracting structured information from dialogues, ensuring visual narrative consistency, and incorporating filmmaking knowledge into the generation process. 

To address these limitations, we propose Dialogue Director, an autonomous framework for dialogue-driven storyboard generation. Dialogue Director leverages large multimodal models (LMMs) without additional training and operates through three specialized agents: the script director, the cinematographer, and the storyboard maker.  
The script director utilizes a Chain-of-Thought reasoning approach \cite{LLMwei2023chainofthoughtpromptingelicitsreasoning} to parse and extract narrative elements such as characters, scenes, and dialogues from input scripts. These extracted elements are organized into a structured database, enabling retrieval-augmented processing \cite{RAG} for subsequent steps.  
The cinematographer generates multi-view references for characters, ensuring consistency in appearance and perspective across shots. It uses retrieved contextual information to maintain alignment between character poses, orientations, and their visual representations in different scenes.  
The storyboard maker integrates cinematic principles into the arrangement of storyboards. It organizes visual elements, such as camera perspectives and character placements, into coherent sequences that align with professional storytelling conventions, ensuring narrative continuity and visual appeal.
Our contributions are as follows:
\begin{itemize}
    \item We introduce a novel task, dialogue visualization, to address the challenges of sparse descriptive details and multimodal complexity in dialogue-driven narratives. 
    \item We firstly propose to use existing metrics to evaluate the new task dialogue visualization, including script understanding and character consistency.
    \item We design a script director to extract and structure narrative elements using Chain-of-Thought reasoning, providing rich context for storyboard generation.
    \item We develop a cinematographer for generating multi-view character representations and a storyboard maker for applying cinematic principles to produce coherent and visually appealing storyboards.
\end{itemize}

\begin{figure*}[t]
  \centering
  \includegraphics[width=0.9\linewidth]{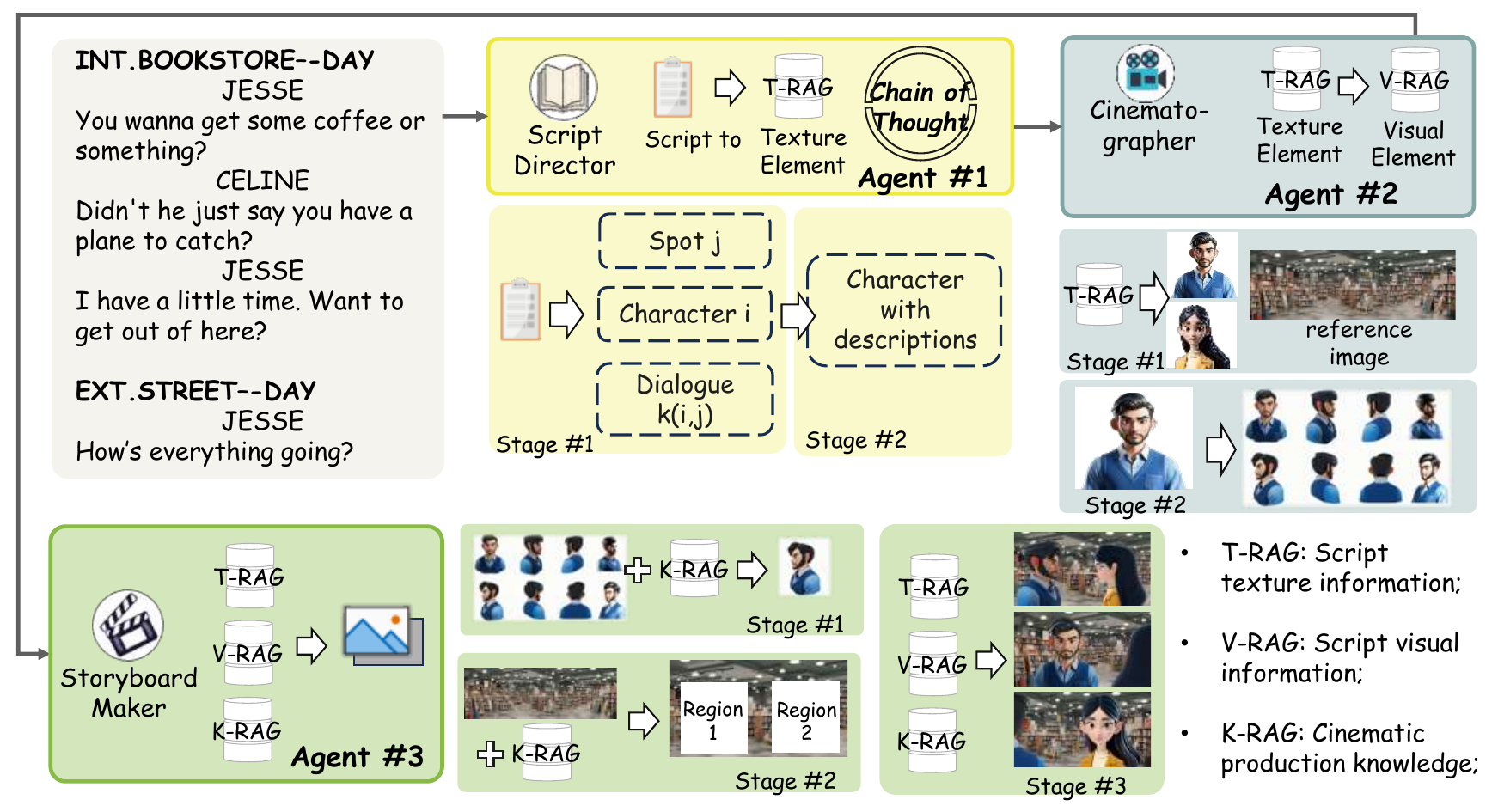}
    \vspace{-0.5cm}
  \caption{Pipeline of our dialogue visualization system, \textbf{Dialogue Director}, consists of three components leveraging LLM for human-like textual understanding. (a) \textbf{Story Director}: Bridges the gap between the dialogue script and detailed scene descriptions for generative models by extracting and enriching key elements. (b) \textbf{Cinematographer}: Visualizes characters and scenes, generating multi-view portraits based on the Story Director's instructions. (c) \textbf{Storyboard Maker}: Combines cinematic knowledge with the dialogue script to plan layouts, select portraits, and compose visual elements into the final storyboard.}
  \label{fig:framework}
\end{figure*}
\section{Related Work}

\subsection{Diffusion Models Leveraging Large Language Models}
Diffusion models \cite{SDrombach2021highresolution,shen2023advancing,gao2024exploring, shen2024imagpose} have shown impressive generative capabilities, with LLMs enhancing their performance through prompt engineering and task decomposition. Recent works leverage LLMs for structured image generation, such as RPG \cite{yang2024mastering} for sub-regional planning, IMAGDressing-v1 \cite{shen2024imagdressing} for editable human image synthesis, and LayoutGPT \cite{feng2024layoutgpt} for layout planning. In diffusion-based storytelling, LLMs refine prompts and plan spatial-temporal regions, as seen in TaleCrafter \cite{gong2023talecrafter} and Anim-Director \cite{li2024anim}. These methods highlight the potential of combining LLMs with diffusion models for structured generation across diverse tasks.
However, they often fail to address the challenges of character consistency and scene diversity in dialogue-driven narratives, crucial for cinematic applications. To bridge this gap, our approach focuses on generating coherent dialogue-driven storyboards directly from narratives or screenplays while maintaining contextual and visual consistency.

\subsection{Story Visualization}
Early works in story visualization, such as StoryGAN \cite{li2018storygan}, primarily utilized GAN or VAE-based methods to generate sequential story images. Subsequent studies aimed to improve visual consistency and multimodal input integration but struggled with maintaining coherence in complex storytelling scenarios. Recently, diffusion models have emerged as a robust alternative, offering controlled and efficient generation.
AR-LDM \cite{arldm} employs a diffusion-based approach for autoregressive story image generation but often introduces artifacts. Rich-contextual Conditional Diffusion Models (RCDMs) \cite{shen2024boosting} enhance semantic and temporal consistency through a two-stage framework, addressing the contextual coherence challenges prevalent in existing methods.
Pre-trained diffusion models with advanced control mechanisms have further improved high-resolution character generation. Techniques utilizing cross-attention, such as StoryMaker \cite{zhou2024storymakerholisticconsistentcharacters}, based on IP-Adapter \cite{CDye2023ip-adapter}, and StoryDiffusion \cite{zhou2024storydiffusionconsistentselfattentionlongrange}, achieve consistent character generation. MIP-Adapter \cite{huang2024resolvingmulticonditionconfusionfinetuningfree} enables multi-character consistency, while In-context-LoRA \cite{lhhuang2024iclorae} excels in storyboard creation and multi-shot video generation. OmniGen \cite{CDxiao2024omnigen} provides fine-grained image control with minimal parameter adjustments.

Despite these advancements, existing methods focus heavily on portrait generation and lack the ability to create conversation-driven scenes with diverse views or extract contextual information from dialogue scripts. This limitation restricts their broader applicability in generative AI for cinematic storytelling. Our approach aims to address these challenges by leveraging dialogue scripts to generate dynamic, multi-view storyboards with contextual and cinematic coherence.

\section{Methods}
\label{sec:methods}

\subsection{Overall}
\label{subsec:overall}
We propose \textbf{Dialogue Director}, a framework that converts a dialogue-centric script $S$ with $D^k$ dialogue segments into a series of cinematic storyboards $V^{K'}$. Dialogue Director fuses language-based reasoning with visual generation methods to produce coherent, multi-view storyboards tailored to dialogue-driven narratives.
Figure~\ref{fig:framework} illustrates three agents in the Dialogue Director pipeline: script analysis and decomposition, multi-view visual generation, and storyboard assembly. The Script Director $\chi$ extracts and refines textual elements, the Cinematographer $\Gamma$ produces multi-view images, and the Storyboard Maker $\zeta$ integrates all components into cinematic layouts. This design ensures that both language-based context and cinematic principles are maintained throughout the entire storyboard creation process.

\subsection{Script Director}
\label{subsec:script_director}

The \textbf{script director} $\chi$ decomposes the script $S$ into characters $C^N$, spots (locations) $P^N$, and dialogues $D^K$. This process leverages a large language model to identify distinct elements that can later be refined and visualized. As shown in Equation~\eqref{eq:extract}, $\chi^{(0)}$ extracts these elements under the guidance of an initial instruction $I^{(0)}$.  
\begin{equation}
\label{eq:extract}
   \Bigl\{\{c^i\}_{i=0}^n,\{p^j\}_{j=0}^n,\{d^l\}_{l=0}^k\Bigr\}
   \;=\;
   \chi^{(0)}\bigl(S, I^{(0)}\bigr).
\end{equation}
Because scripts usually offer limited descriptive information, $\chi^{(1)}$ applies a coarse-to-fine refinement strategy to enrich each extracted element, as indicated in Equation~\eqref{eq:detail}.  
\begin{equation}
\label{eq:detail}
  \{\hat{c}^0,\hat{c}^1,\dots,\hat{c}^n\}
  \;=\;
  \chi^{(1)}\bigl(S, \{c^i\}_{i=0}^n, I^{(1)}\bigr).
\end{equation}
This iterative step ensures that characters and spots have sufficient detail for realistic portrayal in subsequent stages. It also reduces ambiguity when rendering specific appearances or context-related attributes. Finally, the refined data is stored for easy retrieval by the cinematographer, ensuring a consistent flow of information across the pipeline.

% Because dialogues in script $\{d^l\}_{l=0}^k$ have implicit relationship with $\{c^i\}_{i=0}^n,\{p^j\}_{j=0}^n$, $\chi^{(1)}$ leverages the textual understanding of LLM to identify the relationship. Using the $\{c^i\}_{i=0}^n$ and $\{p^j\}_{j=0}^n$, we can form a matrix $X$
% \begin{equation}
% \label{eq:detail}
%   x^{(i,j)}
%   \;=\;
%   \chi^{(2)}\bigl(S, \{d^l\}_{l=0}^k, X, I^{(2)}\bigr).
% \end{equation}

\subsection{Cinematographer}
\label{subsec:cinematographer}

Once the script information is sufficiently detailed, the \textbf{cinematographer} $\Gamma$ generates reference images to capture the visual essence of each character and spot. This step helps transition from purely textual descriptions to visual representations that align with the intended narrative style. Equation~\eqref{eq:refimg} shows how diffusion-based methods~\cite{SDrombach2021highresolution} produce an initial image set:
\begin{equation}
\label{eq:refimg}
    \resizebox{1\linewidth}{!}{$
   \Bigl\{\{cimg^i\}_{i=0}^n,\{pimg^j\}_{j=0}^n\Bigr\}
   \;=\;
   \Gamma^{(0)}\bigl(\{\hat{c}^i\}_{i=0}^n+\theta_{base},\,\{p^j\}_{j=0}^n\bigr),
   $}
\end{equation}
where $\theta_{base}$ denotes baseline prompts (e.g., portrait orientation, framing of the figure). This approach allows for rapid prototyping of visual concepts, which can be iteratively refined. To handle multi-character conversations and varying orientations, Equation~\eqref{eq:multiview} depicts a subsequent step that creates multiple viewpoints $\{cimg^i_x\}_{i=0}^n$ using multi-view diffusion~\cite{huang2024mvadapter,yang2024tencent}:
\begin{equation}
\label{eq:multiview}
   \{cimg^i_x\}_{i=0}^n
   \;=\;
   \Gamma^{(1)}\bigl(\{cimg^i\}_{i=0}^n\bigr),
   \quad
   x \in [0,7].
\end{equation}
These additional perspectives are crucial for depicting realistic character interactions, such as over-the-shoulder shots in dialogue scenes. Generating multi-view images also provides a richer visual library for the storyboard maker, ensuring that the final illustrations appropriately reflect camera angles and character positioning.

\subsection{Storyboard Maker}
\label{subsec:storyboard_maker}

The \textbf{storyboard maker} $\zeta$ synthesizes the script context and multi-view imagery into final storyboard layouts. This stage ensures that linguistic cues and visual elements are merged into coherent, panel-by-panel sequences. Equation~\eqref{eq:viewselect} illustrates how $\zeta$ selects an optimal viewpoint $x^{(i,j)}$ for each dialogue segment $D^{(i,j)}$:
\begin{equation}
\label{eq:viewselect}
   x^{(i,j)}
   \;=\;
   \zeta^{(0)}\bigl(D^{(i,j)}, \{cimg^i_x\}_{i=0}^n, I^{(2)}\bigr).
\end{equation}
Equation~\eqref{eq:boundary} denotes how the storyboard maker assigns layout boundaries $B^{(i,j)}_x$ for positioning characters and scenes:
\begin{equation}
\label{eq:boundary}
   B^{(i,j)}_x
   \;=\;
   \zeta^{(1)}\bigl(D^{(i,j)}, \{cimg^i_x\}_{i=0}^n, \{pimg^j\}_{j=0}^n, I^{(3)}\bigr).
\end{equation}
These boundaries manage character placement and ensure that critical visual elements remain in focus. Finally, Equation~\eqref{eq:final} shows the composition of the final storyboard $V^{K'}$, which combines the selected character views, boundaries, and textual captions:
\begin{equation}
\label{eq:final}
    \resizebox{1\linewidth}{!}{$
   V^{K'}
   \;=\;
   \zeta^{(2)}\Bigl(\{cimg^i_x\}_{i=0}^n, B^{(i,j)}_x, \{pimg^j\}_{j=0}^n, D^{(i,j)}, I^{(4)}\Bigr).
   $}
\end{equation}
This three-agent design combines language reasoning and cinematic composition to create coherent, multi-view storyboards. By integrating metadata like dialogue text and scene notes, it ensures optimal framing and engaging visual storytelling.

\section{Experiments}
We compare DialogueDirector with previous state-of-the-art (SOTA) methods on real-world scripts and narratives for Dialogue Visualization. Empirically, DialogueDirector outperforms existing models in terms of image quality, contextual consistency, and human evaluation metrics. Notably, our method excels in addressing the challenges of understanding real-world dialogue scripts, interpreting the physical world, and incorporating cinematic knowledge, as discussed above.
\noindent \subsection{Experiment settings}
DialogueDirector is designed to be general and extensible, allowing the incorporation of arbitrary MLLMs and multi-view diffusion architectures into the framework. In our experiments, we use GPT-4 \cite{chatgpt4} as the script director and storyboard creator, and employ models based on Stable Diffusion 1.5 \cite{SDrombach2021highresolution} \cite{civitai_model_dream_creation} as our cinematographer to generate reference images. Additionally, we follow the MV-Adapter \cite{huang2024mvadapter} and the first stage of Hunyuan3D-1 \cite{yang2024tencent} as the cinematographer to generate multi-view character portraits. Our implementation runs on a single Nvidia RTX 3090 Ti 24 GB GPU.

\noindent \textbf{Evaluation Metrics. }Selecting evaluation metrics for our novel task, Dialogue Visualization, presents a challenge. We find that CLIP-Image Similarity (CLIP-I) cannot effectively evaluate the side-view appearance of characters. Additionally, while previous research typically uses PSNR (Peak signal-to-noise ratio) or SSIM (Structural Similarity Index) to assess image quality, recent studies have shown that high PSNR or SSIM values do not necessarily correlate with perceived quality by the audience. Therefore, we evaluate the generative quality of various methods using the Natural Image Quality Evaluator (NIQE) \cite{NIQE}. Furthermore, to evaluate the challenges discussed before, we employ CLIP-Text Similarity (CLIP-T) \cite{taited2023CLIPScore} to measure the similarity between descriptive textual information and visual information, assessing both script understanding and visual detail preservation. For evaluating cinematic knowledge integration, we rely on human evaluation.

\noindent \textbf{Baselines.}
It is noteworthy that many existing story-generation methods focus on training on specific datasets, often exhibiting generalization issues with artifacts. Consequently, we perform a comparative analysis of our method against those that demonstrate high controllable generalization capabilities, namely StoryDiffusion \cite{zhou2024storydiffusionconsistentselfattentionlongrange}, OmniGen \cite{CDxiao2024omnigen}, MIP-Adapter \cite{huang2024resolvingmulticonditionconfusionfinetuningfree}, and StoryMaker \cite{zhou2024storymakerholisticconsistentcharacters}. Specifically, StoryDiffusion \cite{zhou2024storydiffusionconsistentselfattentionlongrange} and StoryMaker \cite{zhou2024storymakerholisticconsistentcharacters} focus on storytelling, while MIP-Adapter \cite{huang2024resolvingmulticonditionconfusionfinetuningfree} and OmniGen \cite{CDxiao2024omnigen} specialize in identity-preserving generation. DALL-E 3 \cite{dalle3} demonstrates excellent performance in generating images. Due to their popularity and influence within the community, we position our method as effective in comparison to these models.

\noindent \textbf{Dataset.}
Our framework can process real-world stories and scripts with dialogues, enabling storyboard production without the need for training. In our experiments, we select Hollywood-style scripts rich in dialogue, such as Before Sunset \cite{before_sunset}, as well as stories from The Little Prince \cite{little_prince} and The Little Mermaid from Andersen's Fairy Tales \cite{andersen_fairy_tales}. Due to the limited ability of Large Language Models (LLMs) to handle lengthy texts, we divide the scripts and stories into pages. Each page contains multiple characters engaging in dialogue with scene transitions, demonstrating the complexity of the task. Since most story visualization methods cannot directly process dialogue text, we utilize part of our framework's agents as data processors to extract elements and generate reference images.

\subsection{Quantitative Comparisons}
We evaluate the generated results using the NIQE \cite{NIQE} for image quality and the CLIP-T score\cite{taited2023CLIPScore} for text-image consistency. \tablename~\ref{tab:quantitive} presents the results, with the \textbf{best} scores highlighted in bold and the \underline{second-best} underlined. A lower NIQE score indicates better image quality, as it signifies closer resemblance to natural images. Our method achieves the lowest NIQE score, reflecting superior generative quality. In contrast, StoryDiffusion, which requires additional manual effort and generates fixed-layout comic-style outputs, records the highest NIQE score due to poorer image quality and distortion.
For CLIP-T, a higher score indicates better alignment between textual descriptions and visual details. Our method achieves a competitive CLIP-T score by leveraging collaboration between its three agents, ensuring coherent integration of textual and visual information. Unlike existing methods, which demand extra manual preprocessing, fail to process dialogue scripts, and struggle with multi-view coherence or cinematic layouts, our approach overcomes these limitations, providing an efficient and flexible solution with superior quality and coherence.

\begin{table}[t]
    \centering
      \vspace{-0.5cm}
        \caption{Quantitive camparisons with state-of-the-art methods, NIQE weights the quality of image generation. Clip-T weights the coherence with the visual information.
    }
            \setlength\tabcolsep{6pt}
    \resizebox{1\linewidth}{!}{%
    \begin{tabular}{lccccc}
        \toprule
  \textbf{Approach} & \makecell{\textbf{Without extra}\\{ \textbf{Manual effort}} } & \makecell{\textbf{Detail}\\\textbf{Coherence}} & \makecell{\textbf{Cinematic}\\\textbf{Layout}} & \makecell{\textbf{NIQE\cite{NIQE}↓}} & \makecell{\textbf{CLIP-T\cite{taited2023CLIPScore}↑}} \\
        \midrule
     MIP-Adapter\cite{huang2024resolvingmulticonditionconfusionfinetuningfree} & \xmark & \xmark & \xmark & 5.20 & 0.1543\\

    StoryMaker\cite{zhou2024storymakerholisticconsistentcharacters} & \xmark & \xmark & \xmark & \underline{3.91} & 0.1980\\

  StoryDiffusion\cite{zhou2024storydiffusionconsistentselfattentionlongrange} & \xmark & \xmark & \xmark & 5.32 & \textbf{0.2247} \\

  \textbf{Ours} & \cmark & \cmark & \cmark & \textbf{3.78} & \underline{0.2240}\\
        \bottomrule
        \vspace{-1.75em}
    \end{tabular}
    }
    %\vspace{-0.5cm}
    \label{tab:quantitive}
\end{table}

\begin{figure}[t]
  \centering
  \includegraphics[width=1.0\linewidth]{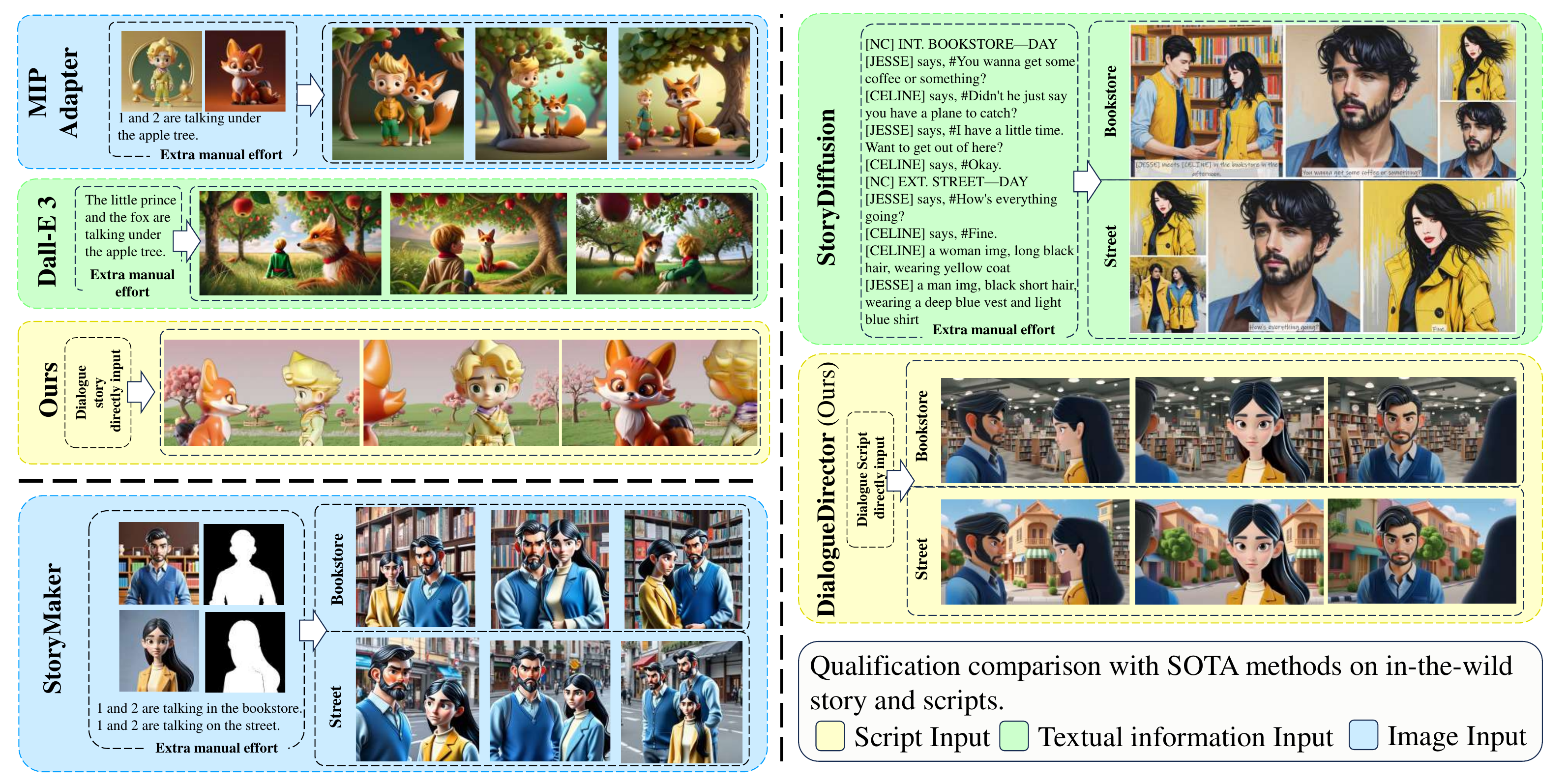}
%    \vspace{-0.5cm}
  \caption{In-the-wild scripts Dialogue Visualization evaluation. Methods in yellow frame mean the use of script only, without any manual effort; methods in green frame mean use textual information with manual effort; methods in blue frame mean the use of reference images generated by the agent cinematographer.}
  % \vspace{-0.5cm}
  \label{fig:exp}
\end{figure}

\begin{table}[t]
    \centering
    \vspace{-0.5cm}
        \caption{Human evaluation with existing storytelling methods on three metrics: The storyboard should reflect Complex relationship, maintain details in different views and apply cinema of knowledge properly.
    }
            \setlength\tabcolsep{6pt}
    \resizebox{1\linewidth}{!}{%
    \begin{tabular}{lcccc}
        \toprule
  \textbf{Approach} & \makecell{\textbf{Use Dialogue}\\{ \textbf{Script Only}} } & \makecell{\textbf{Complex}\\{\textbf{Relationship↑}}}& \makecell{\textbf{Physical}\\{\textbf{understanding↑}}}& \makecell{\textbf{Cinema of }\\{\textbf{knowledge↑}}}\\
        \midrule
    OmniGen(text)\cite{CDxiao2024omnigen} & \cmark  & 2.50 & 2.10 & 2.57 \\

  OmniGen(image)\cite{CDxiao2024omnigen} & \xmark & 2.97 & \underline{3.17} & \underline{2.97} \\

     MIP-Adapter\cite{huang2024resolvingmulticonditionconfusionfinetuningfree} & \xmark & 2.67 & 2.13 & 2.67\\

    StoryMaker\cite{zhou2024storymakerholisticconsistentcharacters} & \xmark & 2.27 & 2.00 & 2.20\\

  StoryDiffusion\cite{zhou2024storydiffusionconsistentselfattentionlongrange} & \xmark & \underline{3.03} & 2.90 & 2.90 \\

  \textbf{Ours} & \textbf{\cmark}  & \textbf{3.67} & \textbf{4.00} & \textbf{3.47}\\

        \bottomrule
        \vspace{-1.75em}
    \end{tabular}
    }
    \label{tab:human}
\end{table}

\subsection{Qualitative Comparisons}
We use the best results from SOTA methods, noting that diffusion-based approaches cannot consistently generate identical portraits. Some methods also require external tools for text pre-processing or reference images, whereas our Dialogue Director handles these tasks independently and can even assist other methods.
As shown in \figurename~\ref{fig:exp}, while SOTA methods excel at maintaining character appearance in portraits, they struggle to depict stories with dialogues and maintain character details across views. MIP-Adapter and StoryMaker fail to interpret shot changes in conversations: MIP-Adapter preserves facial features but loses clothing details, while StoryMaker requires pre-processed masks and inconsistently portrays characters across angles, often altering hairstyles or clothing. DALL-E 3 understands cinematic principles but fails to maintain character details. StoryDiffusion, despite extra manual input, cannot ensure consistent multi-view character representation while applying cinematic knowledge. In contrast, our approach addresses these challenges effectively, demonstrating the strength of our framework.

\subsection{Human evaluation}
30 college students majored in film were invited to evaluate these methods. Through 5 pieces in-the-wild script Dialogue Visualization with multiple characters and spot change, users are asked to rate the results on a Likert scale of 1 to 5, according to the following criteria:  \textbf{(i) Complex Relationship Reflect script.} Generation results should reflect the conversation relationship, including people's eye contact and face orientation. \textbf{(ii) Physical understanding}. Characters and decoration details should maintain consistency during shot change and orientation change. \textbf{(iii) Cinema of knowledge. }The results should be compatible with the dialogue, adhering to cinematic principles. A higher score indicates better performance. The average scores in the three aspects are shown in \tablename~\ref{tab:human}. The \textbf{best} results are shown in bold, and the \underline{second best} are shown in underline. Our approach outperforms the second-best by a large margin, highlighting the effectiveness of our proposed framework.

\begin{table}[t]
    \centering
        \caption{Ablation analysis. We make ablation analysis on our method on NIQE and Clip-T. Noticing that our method is plug-and-play, we use it on OmniGen to compare the results.
    }
    \setlength\tabcolsep{6pt}
    \resizebox{1.0\linewidth}{!}{%
    \begin{tabular}{lccccc}
        \toprule
  \textbf{Approach} & \makecell{\textbf{Without extra}\\{ \textbf{Manual effort}} } & \makecell{\textbf{Detail}\\\textbf{Coherence}} & \makecell{\textbf{Cinematic}\\\textbf{Layout}} & \makecell{\textbf{NIQE\cite{NIQE}↓}} & \makecell{\textbf{CLIP-T\cite{taited2023CLIPScore}↑}} \\
        \midrule
  OmniGen \cite{CDxiao2024omnigen} (w/text) & \cmark & \xmark & \xmark & 4.86  & 0.1442  \\
  OmniGen \cite{CDxiao2024omnigen} (w/image) & \cmark & \cmark & \xmark & \textbf{3.62} & 0.1867  \\
     w/o Script Director & \xmark & \xmark & \xmark & 3.96 & 0.1419 \\
         w/o Cinematographer & \xmark & \xmark & \xmark & 4.51 & 0.1332\\
           w/o Storyboard Maker & \xmark & \cmark & \xmark & 4.09 & \underline{0.2096} \\
             \textbf{Ours} & \cmark & \cmark & \cmark & \underline{3.78}  & \textbf{0.2240} \\
        \bottomrule
        \vspace{-1.75em}
    \end{tabular}
    }
    % \vspace{-0.5cm}
    \label{tab:ablation}
\end{table}

\begin{figure}[t]
  \centering
  \includegraphics[width=1.0\linewidth]{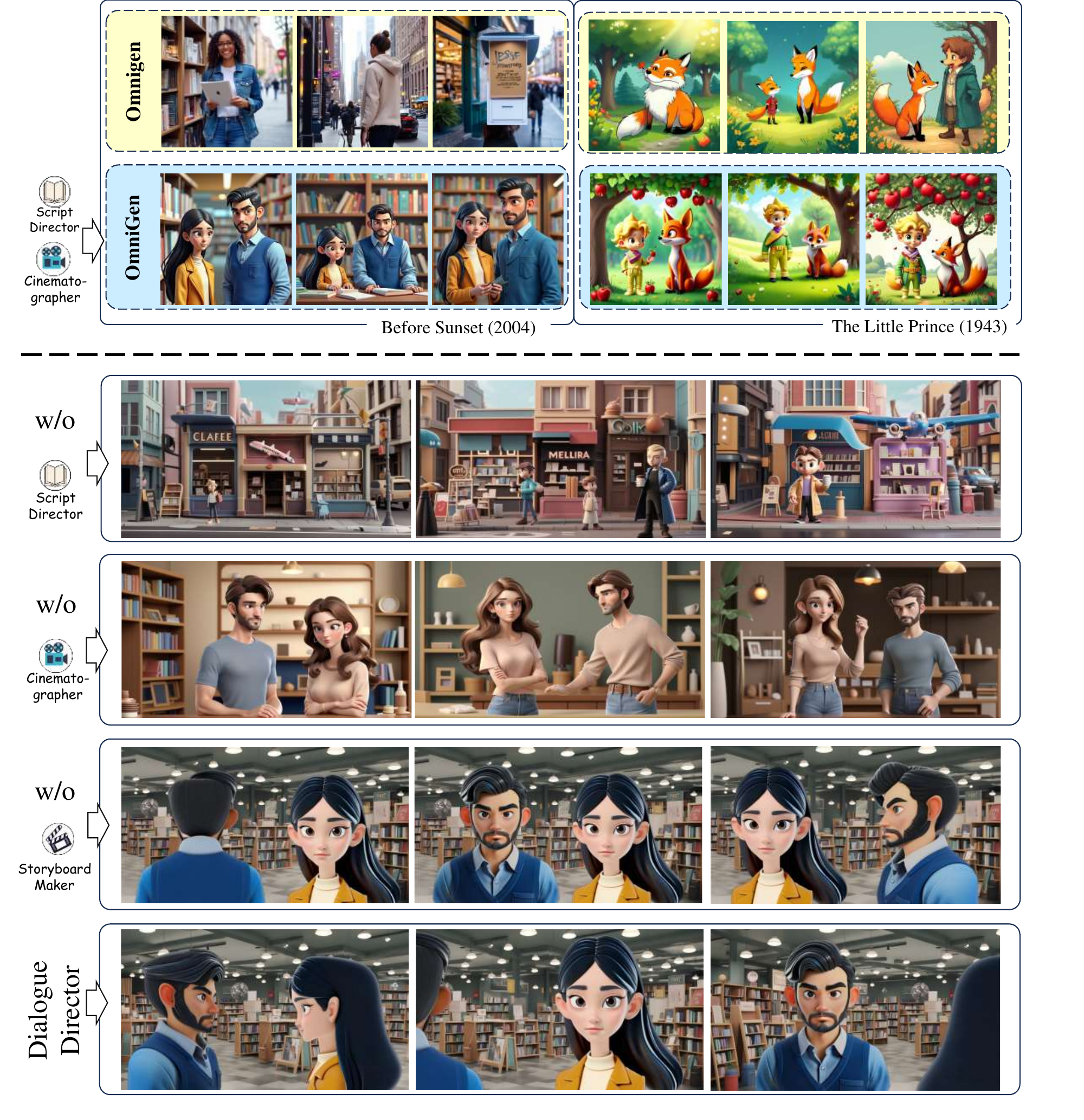}
  \vspace{-0.5cm}
  \caption{Ablation analysis on our method and OmniGen. It can be seen the agents in our method can act as plug-and-play component in other generative method like OmniGen. The three components perform their duties as we expect.}
  \label{fig:ablation}
    \vspace{-0.5cm}
\end{figure}

\subsection{Ablation Analysis}

Our Dialogue Director is general and extensible, allowing us to incorporate these three components into existing state-of-the-art (SOTA) image generation architectures. OmniGen \cite{CDxiao2024omnigen}, a powerful method capable of processing both textual and visual information, performs poorly when directly handling scripts, as demonstrated by the generation results in \figurename~\ref{fig:ablation}. Like other existing generative methods, it struggles to extract meaningful visual information from dialogues. This leads to suboptimal performance in both the NIQE and CLIP-T evaluations. Our method addresses these limitations by helping OmniGen extract visual information from dialogues and providing reference images to maintain consistent details. The significant improvement in performance highlights the efficacy of our approach.

\noindent \textbf{Effect of Script Director.} We leverage the script director to effectively understand in-the-wild scripts and narrative stories. As shown in \figurename~\ref{fig:ablation} and \tablename~\ref{tab:ablation}, without the script director, methods fail to capture implicit information in dialogues, resulting in irrelevant storyboards for characters and scene transitions.

\noindent \textbf{Effect of Cinematographer.} The cinematographer ensures that characters and scene details remain consistent across storyboards. As illustrated in \figurename~\ref{fig:ablation} and \tablename~\ref{tab:ablation}, without the cinematographer, the generated results lack visual coherence and fail to maintain continuity in the story during shot transitions.

\noindent \textbf{Effect of Storyboard Maker.} As shown in \figurename~\ref{fig:ablation}, our method produces a well-structured layout with cinematic knowledge, thanks to the storyboard maker. Without this component, the generated results exhibit unrealistic layouts and illogical character orientations, which is also reflected in the NIQE and CLIP-T metrics.

\section{Conclusion and Limitations}
In this paper, to address the challenges of generating dialogue storyboards from dialogue scripts in practice, we propose Dialogue Visualization task, and further a training-free framework called \textbf{Dialogue Director}, which leverages Multimodal Large Language Models (MLLMs) to enhance vivid story-telling. In Dialogue Director, we employ multi-view diffusion models as the Cinematographer, working in tandem with our MLLM-based Script Director and Storyboard Maker. This collaborative approach addresses the key challenges encountered in Dialogue Visualization. Extensive qualitative and quantitative experiments demonstrate the effectiveness of \textbf{Dialogue Director}. As for limitations, the modeling performance relies on the characters’ multi-view closely. Also, storyboards with dynamic shots and characters' complex pose would be difficult to generate, which will be treated as our future work.

% \section*{References}

\bibliographystyle{ieeetr}
\bibliography{main}

\end{document}